# A Hybrid, PDE-ODE Control Strategy for Intercepting an Intelligent, well-informed Target in a Stationary, Cluttered Environment

Ahmad A. Masoud
Electrical Engineering Department, KFUPM, P.O. Box 287, Dhahran 31261, Saudi Arabia,
e-mail: masoud@kfupm.edu.sa

**Abstract**

In [1,2] a new class of intelligent controllers that can semantically embed an agent in a spatial context constraining its behavior in a goal-oriented manner was suggested. A controller of such a class can guide an agent in a stationary unknown environment to a fixed target zone along an obstacle-free trajectory. Here, an extension is suggested that would enable the interception of an intelligent target that is maneuvering to evade capture amidst stationary clutter (i.e. the target zone is moving). This is achieved by forcing the differential properties of the potential field used to induce the control action to satisfy the wave equation. Background of the problem, theoretical developments, as well as, proofs of the ability of the modified control to intercept the target along an obstacle-free trajectory are supplied. Simulation results are also provided.

## I. Introduction:

Agents are built using taskable processes capable of exhibiting a yielding, purposive, intelligent behavior. The activities in a selected process are required to coalesce in order to produce a desired function of some sort. There are two opposing views regarding the identity of a function an agent may acquire: the first view stresses that a function can be made intrinsic to an agent where the agent continues to perform the same function that is encoded in it irrespective of the situation it operates in. The other view, which significantly differs from the first one, stresses that the assets under the disposal of an agent and the manner they are being utilized are not enough on their own to determine what the agent's function is. Context, with which the environment presents the agent, is also a major factor in determining the identity of an agent's function. A change in context will result in a change in function even if the assets under the disposal of the agent and the manner in which they are being deployed are not changed.

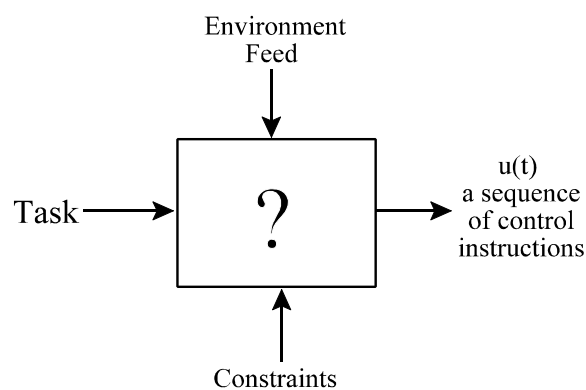

Figure-1: A planner is essential for the functioning of an agent.

Sensitivity to context is what makes the behavior of an agent useful and meaningful. Such type of functionality is realized by a device called a planner. A planner is a goal-oriented, context-sensitive, intelligent controller. It functions to combine the sensory feed from an environment, constraints on behavior, specified task, and agent's model to generate a sequence of action instructions the agent may use to deploy its actuators of motion in order to project a behavior that actualizes the function the agent is required to perform (figure-1).

Designing a planner for an agent is a challenging task. There are many issues a planner has to deal with if the utilizing agent is to have a reasonable chance of success performing a function in a realistic environment. Some of these issues are: coping with ambiguity and incomplete information about the environment, timely generation of the navigation control signal, managing the agent's dynamics and placing limitations on the manner it can project motion, etc. The focus of this paper is on context. In particular, we consider the situation where the context in which the agent is operating attempts to impede its ability to function.

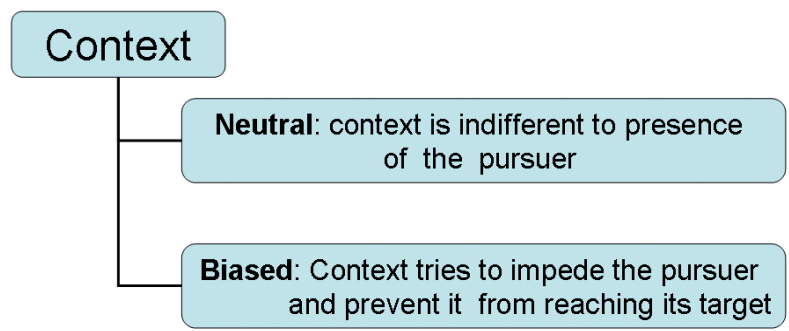

Figure-2: Context may be divided into neutral and biased.

Context may be divided into two types: neutral and biased (figure-2). A neutral context is indifferent to the presence of the agent and the type of activities it engages in. On the other hand, a biased context attempts to impede an agent and prevent it from carrying out the intended function.

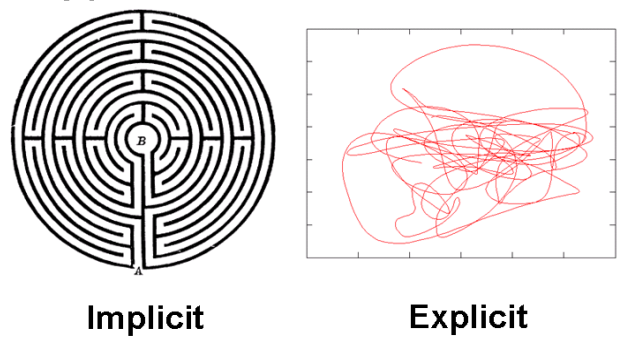

Figure-3: biased context may be divided into implicit and explicit.

Biased context may be divided into two types: implicit and explicit (figure-3). Implicit bias is subtle. It is encoded in the structure of the environment and uses a model of the psychology of the pursuer in order to confuse it and guide it away from the target. An example of this type of bias is a maze carefully designed to be devoid of any cues or structural regularities that are discernable by the pursuer. The aim is to bewilder it and steer it away from the static target [3,4,5]. On the other hand, an explicitly biased environment overtly moves the goal away from the pursuer each time it is about to be reached. In this case the goal is called an evader. It is also possible to have a mixed biased that is a hybrid of the two.

Explicit bias may be divided into two: active and passive (figure-4). In the passive case the evader is not informed about the movements of the pursuer; it is only aware of its presence. It attempts to evade capture by blindly executing an escape maneuver. On the other hand, an active evader is well informed about the movements of the target even its intentions. Its escape maneuver keeps adapting to the knowledge base it has about the pursuer.

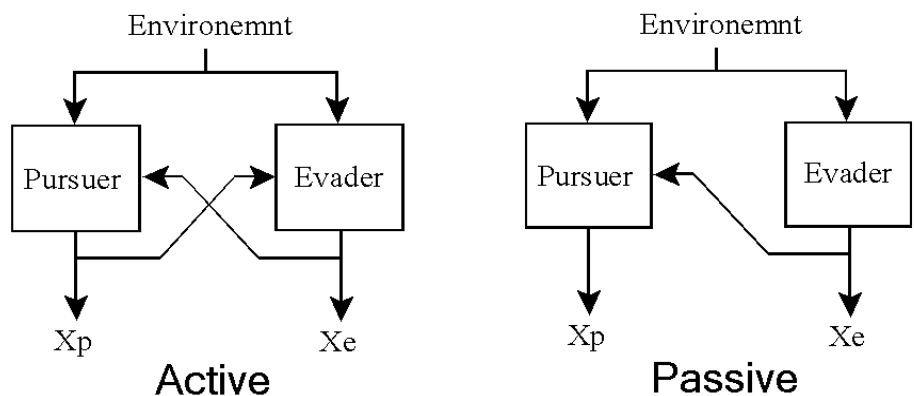

Figure-4: explicit bias may be divided into active and passive.

The interest in this paper is in the design of a planner for a pursuer that is working in a biased, mixed, active context. The planner is used for attempting to capture an intelligent, well-informed evader that is actively maneuvering to evade capture in a cluttered environment utilizing both advantages of the terrain and deficiencies in the pursuer's decision making process. This type of pursuit-evasion problem is important. The survival of a species is dependent, upon other things, on its ability to deal with such a type of problem, i.e. develop its navigation competence to successfully handle a prey-hunter situation. The developed capabilities are usually subjective in nature, that is the structure of the capture scheme is built around cues that are meaningful to the hunter and related to aspects of the prey's personality. In a situation where an intelligent prey is being hunted, the prey may be aware of the hunter. Even more, it may be aware of what the hunter expects from it. In such a situation, the prey may initiate a deception scheme that is masked by its expected behavior with the goal of engaging in an interactive message exchange with the hunter. The structure of this exchange is based on the model which the prey has for the hunter's decision making process. Its aim is to out-maneuver its pursuer and evade capture. Through this exchange, the prey may even acquire a soft-control of the hunter that could reverse the role of each. Here, the nesting of action release, or equivalently message exchange, has the form of I KNOW that IT KNOWS that I KNOW etc. [6]. The deeper the nesting is, the more likely that the actions released by the party concerned drive it to the desired conclusion.

Such subjectivity seems inherent in the structure of these kind of problems. This has convinced the majority of researchers in the field of the need to use evolutionary techniques (e.g. neural networks, genetic algorithms, etc.) to "absorb" the personality profile of the target, hence creating a predictor of behavior, in order to derive a successful capture scheme [7,8]. While profiling may work most of the time, the situation is dramatically different for the case of active intelligent targets. These targets are capable of adjusting their trajectories to enhance their chance of survival by extending their domain of awareness to include that of their pursuers. To the best of this author's knowledge, this class of problems has not been addressed in the literature [9].

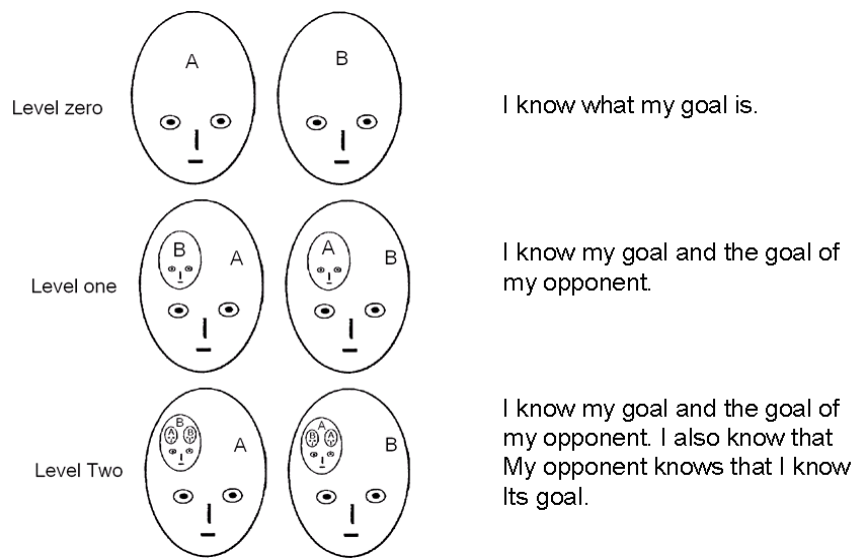

Figure-5: Nested awareness in a dual

This paper hypothesizes the existence of objective, provably-correct and profile-free techniques for target interception. The ability of such techniques to exert a successful planning action is directly tied to the potential of achieving a desired conclusion instead of being dependant on the psychology of the prey. Testing the above hypothesis is carried out with the help of a newly introduced class of intelligent motion controllers called: hybrid, partial differential equation - ordinary differential equation (PDE-ODE) motion planners [1,2]. Untill now such controllers have been implemented using harmonic potential fields (HPFs). HPFs possess many useful properties that make them excellent tools for navigation [10]. Most notably, such types of planners guarantee, irrespective of the geometry or topology of the context, that a path to the target will be found if it exists, otherwise the planner will give an indicator that the problem is unsolvable and there is no path to the target.

Despite the efficiency of HPF-based methods in tackling environments with sophisticated geometry, they can only engage simple, stationary targets (a sitting duck). In this paper a modified version of the control that utilizes the wave potential is suggested for engaging active, intelligent targets in a provably-correct manner.

The rest of the paper is organized as follows: section II provides a quick background of PDE-ODE planners. In section III the suggested planner is presented. Proofs of performance are given in section VI. Simulation results and conclusions are placed in sections VII and VIII respectively.

## II. PDE-ODE Planners: A Background

The backbone of any planner is an action selection mechanism. This mechanism is responsible for generating a sequence of instructions $\{u_0,...,u_L\}$ which, when fed to the agent's actuators of motion, yield a sequence of state transitions $\{X_0,...,X_L\}$ connecting an initial point to a target point so that the final state $X_L$ is the goal state of the agent, and all the transient states satisfy the constraints on behavior. This sequence is called a plan. In a PDE-ODE planner (figure-6), this plan is a member of a field of plans called the action field.

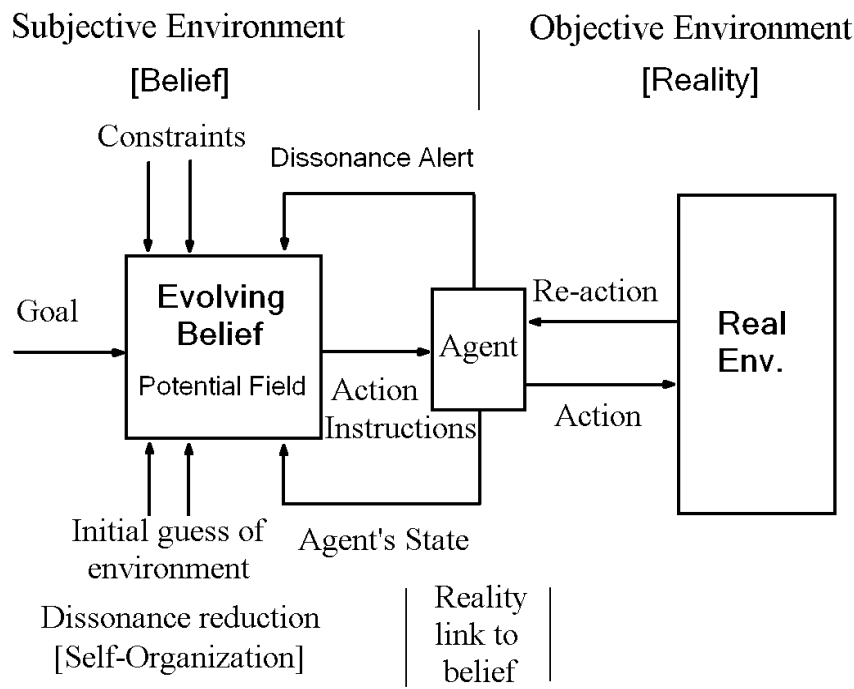

Figure-6: A PDE-ODE Controller.

The action field assigns to each possible situation the agent is expected to encounter an action instruction that is supposed to make it move closer to its target. The planner exhibits a negative entropy effect to reduce the ambiguity and disorder caused by the large number of possible actions and action combinations the agent can choose from. It unifies all the differential actions in one structure that may be used as an action field by the agent (figure-7).

There are two major approaches for action selection which an agent may use for generating a plan. A plan is either generated on the basis of an intelligent (or brute force) search of all possible action values and combinations, or a plan is evolved. Search-based methods were the first to be used for such a purpose. It was observed that the performance of search-based techniques in an experimental setting are significantly inferior to that predicted by simulation. The chances that an agent may fail or at best have a shaky and precarious response are significant. Brooks studied the causes of such discrepancies [11] and suggested four conditions a planner has to satisfy in order for the unitizing agent to have a reasonable chance of success operating in a realistic environment. He found that a successful planning action must be embodied, situated, intelligent and emergent. In general, evolution-based techniques are found to better suit these four requirements.

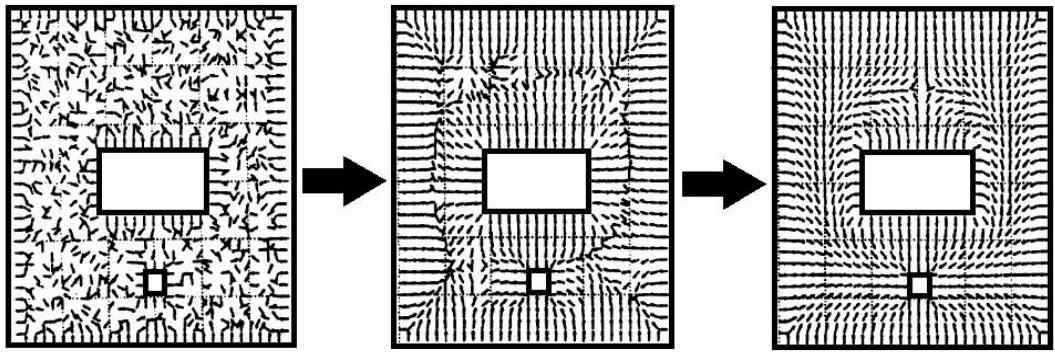

Figure-7: Evolution of action field in a PDE-ODE planner.

PDE-ODE intelligent controllers (planners) use evolution to synthesize an action field. The field is generated by the synergetic interaction of a massive number of differential systems (micro-agents), figure-8. A micro-agent has a form that is identical to the agent being controlled, with the exception that its state is stagnant and immobilized to an *a priori* known location in state space.

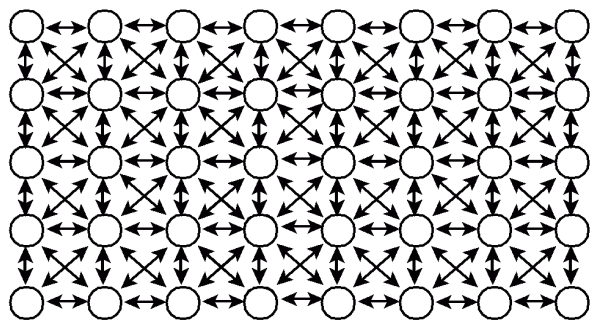

Figure-8: A collective of micro-agents.

The actions of the micro-agent collective are emulated using a potential field that is operated on by a vector partial differential operator. The synergetic interaction needed to convert the individual actions of the micro-agents into a guidance-capable group action is achieved in the following two steps:

1- the individual micro-agents are informationally coupled (figure-9). This is achieved using the proper partial differential governing relation,
2- the process of morphogenesis [12], which is responsible for guiding the group structure of the action field, is activated by factoring the influence of the agent's environment in the behavior generation process. This influence is factored-in using boundary control action. The overall control structure is induced by solving the boundary value problem (BVP) that is constructed using the partial differential governing relation from step-1, and the boundary control action.

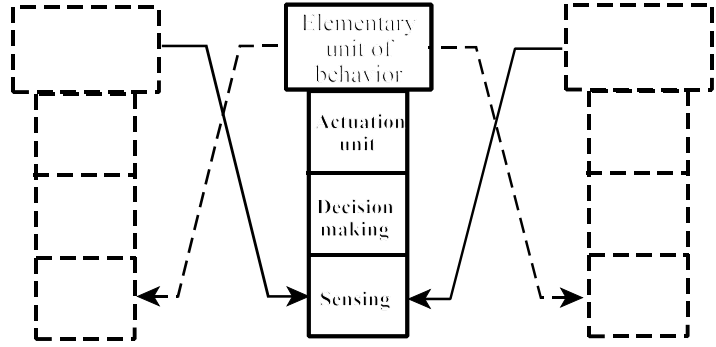

Figure-9: An informationally-coupled micro-agent.

The above mode of behavior synthesis is in conformity with the artificial life approach to behavior generation [13].

### III. The Suggested PDE-ODE Planner

Planning the movements of a pursuer in a manner that can cope with an active, intelligent evader is carried out using the nonlinear, dynamical system:

$$\dot{x} = g(V(x(t), x_p(t), \Gamma)) \qquad x(0) \in \Omega. \qquad 1$$

The above system is required to make:

$$\lim_{t \to \infty} |x(t) - x_p(t)| \to 0, \qquad 2$$

and
$$x(t) \in \Omega \quad \forall \, t,$$

where $\Omega$ is an admissible subset of the N-dimensional state space, $\Gamma$ is the boundary of the non-admissible subset of state space (O, $\Gamma = \partial O$), $x_p(t)$ is the trajectory of the target ($x_p(t) \in \Omega \ \forall \, t$), $x(t) \in R^N$, g: $R \to R^N$, V: $R^N \times R^N \times R^M \to R$, where $\Gamma \in R^M$, $M \leq N-1$. The above implies the following:

1- The target may be intelligent,
2- The target may have an accurate model of its environment, as well as information about the movements of its pursuer,
3- No psychological profile of the target, its tendencies, and habits or, for that matter, a statistical model of the target's behavior are needed,
4- The pursuer has a good model of the environment, and full information about the movements of the target.

The target is assumed to have limited power so that it cannot instantly change its position or orientation (i.e. $x_p(t) \in C^L$, $L \geq 2$).

The planning action from the nonlinear dynamical system in (1) is derived at two stages. A potential field that covers the space of all possible situations (states) the agent may encounter is first derived. A properly chosen differential operator is then used to extract the action the agent needs to execute at a certain situation. The potential field is the backbone of a PDE-ODE planner. It is supposed to be configured in a manner that enables it to emulate the AL machine discussed in section II. There are many operations a potential field-based, virtual, intelligent, AL machine has to perform in order to provide the guidance function an agent needs in order to reach its goal. For example, the potential must tie the spatial and temporal variations of the situation it is dealing with together in order to generate the implicit prediction capabilities the pursuer needs to counteract the movements of the evader. It must also hard-encode the actions the agent needs to take when it encounters certain components of the environment. These actions are derived based on *a priori* determined survival behavior the agent must exhibit when such components are faced. The main operation the potential field has to carry out is to generate the structure for the goal-oriented group from which the action field is constructed. The construction has to be carried out in a manner that harmoniously encode all the desired aspects of behavior.

The field generates and stores in the differential properties of the potential all action-situation pairs that are potentially usable by the agent during the execution of its function. The second stage used by a PDE-ODE planner uses a differential operator to tap into the stored potential actions and actualizes the one designated for the situation the agent is currently encountering. In the following sections the two stages used by the suggested planners are presented.

**A. The PDE Component:**
Scalar potential fields that describe changes in both space and time are suitable for synthesizing action fields that can be used for tracking moving targets. The partial differential relation that may be used to govern the differential properties of such surfaces is the wave equation (WE):

$$\nabla^2 V = \frac{1}{a^2} \frac{\partial^2 V}{\partial t^2} \qquad 3$$

where a is a positive constant, $\nabla$ is the gradient operator, $\nabla^2$ is Laplace operator. The reason the WE is chosen over other spatio-temporal governing relations (e.g the diffusion equation) is mainly due to the nature of its solution. From the method of separation of variables, the solution of a field that is dependent on both space and time ($V(x,t)$), and is governed by the WE may be written in the form:

$$V(x,t) = R(x)T(t) \qquad 4$$

where R is the position-only dependent component of the solution, and T is the time-only dependent component. Therefore, the WE may be placed in the form:

$$T \cdot \nabla^2 R - \frac{1}{a^2} R \frac{\partial^2 T}{\partial t^2} = 0. \qquad 5$$

The only way equation 5 can be satisfied is for the position and time dependent terms to be equal to the same constant which is, for convenience, chosen equal to $-\lambda^2$:

$$\frac{\nabla^2 R}{R} = a^2 \frac{\partial^2 T / \partial t^2}{T} = -\lambda^2 \qquad 6$$

As a result, T(t), and R(x) may be computed by solving the following Helmholtz equations (HEs):

$$\nabla^2 R + \lambda^2 R = 0 \qquad \text{N-D HE} \qquad 7$$

$$\frac{\partial^2 T}{\partial t^2} + (a\lambda)^2 T = 0 \qquad \text{1-D HE}$$

It is well-known that the fundamental solution of an N-D HE provides N orthogonal basis functions capable of representing an arbitrary scalar function that is defined on that space. Therefore, the above set of equations yields N+1 orthogonal basis enough to represent (using the generalized Fourier series expansion) any piecewise continuous function in both time and space.

A. *BVP-1, The Dirichlet Case:*

The generating BVP is: solve $\quad \nabla^2 V = \dfrac{1}{a^2}\dfrac{\partial^2 V}{\partial t^2}. \qquad 8$

subject to $V(x,x_p(t),\Gamma)=C$ for $x\in\Gamma$, and $V(x_p(t),x_p(t),\Gamma)=0$, where C is a positive constant, and $x_p(t)$ is the target's trajectory (figure-10).

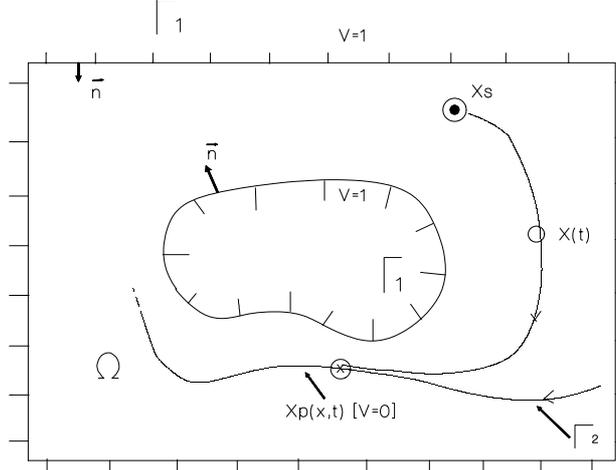

Figure-10: Boundary Conditions, the Dirichlet case.

B. *BVP-2, The Homogeneous Neumann Case:*

The generating BVP is: solve $\quad \nabla^2 V = \dfrac{1}{a^2}\dfrac{\partial^2 V}{\partial t^2}. \qquad 9$

subject to $V(x(0),x_p(t),\Gamma)=C$, $\partial V(x,x_p(t),\Gamma)/\partial n=0$ for all $x\in\Gamma$, and $V(x_p(t),x_p(t),\Gamma)=0$, where **n** is a unit vector normal to $\Gamma$. Existence and uniqueness of the solution of the above BVPs were proven in [14,15].

**B. The ODE System**:
The interception trajectory is generated using the first order nonlinear differential system:

$$\dot{x} = g(x(t), x_p(t), \Gamma) \qquad 10$$

$$g(x,x_p,\Gamma) = -[\nabla V(x(t),x_p(t),\Gamma) + \frac{\nabla V(x(t),x_p(t),\Gamma)}{\|\nabla V(x(t),x_p(t),\Gamma)\|^2}\frac{\partial V(x(t),x_p(t),\Gamma)}{\partial t}]$$

It can be shown that the above ODE system is capable of satisfying condition 2. Unfortunately, the system encounters a singularity when $x(t)=x_p(t)$. To remedy this problem, the condition in 2 is relaxed to:

$$\lim_{t \to \infty} |x(t) - x_p(t)| \leq \rho, \qquad 11$$

where $1 >> \rho > 0$. A singularity-free ODE system that can accomplish the above may be used:

$$g(x, x_p, \Gamma) = -[\nabla V(x(t), x_p(t), \Gamma) + \frac{\nabla V(x(t), x_p(t), \Gamma)}{\beta(\|\nabla V(x(t), x_p(t), \Gamma)\|^2)} \frac{\partial V(x(t), x_p(t), \Gamma)}{\partial t}]$$

where
$$\beta(x) = \begin{bmatrix} x & x \geq \rho \\ \eta(x) & x < \rho \end{bmatrix} \qquad 12$$

and $\eta(x)$ is a monotonically increasing function that satisfies the following conditions:

$$\eta(0) = \epsilon, \quad \eta(\rho) = \rho \qquad \rho > \epsilon > 0,$$

$$\frac{d\eta(x)}{dx}\bigg|_{x=\rho} = 1, \quad \frac{d\eta(x)}{dx}\bigg|_{x=0} = 0. \qquad 13$$

A form for $\eta(x)$ that satisfies the above conditions is:

$$\eta(x) = \varepsilon + \frac{2\rho - 3\varepsilon}{\rho^2} x^2 + \frac{2\varepsilon - \rho}{\rho^3} x^3. \qquad 14$$

It ought to be noticed that the partial, implicit dependence of $\partial V/\partial t$ on $dx_p/dt$ does not imply that the velocity of the target is needed for its computation. Any numerical procedure for the solution of the time dependent potential computes $\partial V/\partial t$ using the a finite-difference approximation:

$$\frac{V(x(t), x_p(t), \Gamma) - V(x(t-dt), x_p(t-dt), \Gamma)}{dt}. \qquad 15$$

Constructing the above approximation requires only that the easier to compute, time-dependant position of the target be estimated.

A. *The solution for the ODE System in 12 Exists*
The nonlinear ODE system in 12 satisfies the global Lipschitz condition [16]. In other words, for every time interval $\tau \in [0, \infty)$, $\exists$ the constants $k\tau < \infty$, and $h\tau < \infty$, so that

$$\|g(x, x_p(t), \Gamma) - g(y, x_p(t), \Gamma)\| \leq k\tau \|x - y\| \qquad 16$$

and
$$\|g(x(0), x_p(t), \Gamma)\| \leq h\tau,$$

$\forall t \in [0, \tau]$, and $\forall x, y \in R^n$. This means that the solution of 12 does exist, and is unique. This in turn implies:
1- $x$ is differentiable almost everywhere (i.e. $dx/dt$ exists),
2- the relation: $\qquad \dot{x} = g(x(t), x_p(t), \Gamma) \qquad 17$
holds for t where $dx/dt$ exists,
3- $x(t)$ satisfies: $\qquad x(t) = x(0) + \int_0^t g(x(\tau), x_p(\tau), \Gamma) d\tau. \qquad 18$

## VI. Motion Analysis

In this section, the ability of the suggested PDE-ODE system to enforce the convergence and avoidance constraints in 2 is examined. The proof is based on Liapunov second method [17]. A key element of the proof is showing that the wave potential in section IV is a Liapunov function candidate (LFC).

### A. The Wave Potential is an LFC

It is well-known that the solution of the WE is analytic. This satisfies the requirement that a LFC be differentiable, or at least continuous. It can also be shown that surfaces with differential properties that are governed by the linear, elliptic, WE partial differential operator satisfy the maximum Principle (i.e they are free of local extreme, where minima or maxima of such functions can only occur on the boundary of the space on which V is defined ($\Gamma$)), [18]. This in turn leads to the satisfaction of the second condition required by an LFC, that is:

1- $V(x(t_i), x_p(t_i), \Gamma) = 0$     at, and only at $x=x_p$,            19
2- $V(x(t_i), x_p(t_i), \Gamma) > 0$      $x \in \Omega$, $\forall t_i \in t$.

### B. Liapunov's Direct Method

A point $x_p$ is considered to be an equilibrium point of the system in 1 if

$$g(x_p(t), x_p(t), \Gamma) = 0 \qquad \forall t \geq 0 . \qquad 20$$

This equilibrium point is considered to be globally, asymptotically stable (i.e. $x(t) \to x_p(t)$ as $t \to \infty$) if $\exists$ a LFC $\ni$:

1- $dV(x(t_i), x_p(t_i), \Gamma)/dt = 0$     at and only at $x=x_p$,         21
2- $dV(x(t_i), x_p(t_i), \Gamma)/dt < 0$     $x \in \Omega$, $\forall t_i \in t$.

### C. Convergence Analysis

The time derivative of the wave potential is:

$$\dot{V}(x(t), x_p(t), \Gamma) = \nabla V(x(t), x_p(t), \Gamma)^t \dot{x} + \frac{\partial V(x(t), x_p(t), \Gamma)}{\partial t}. \qquad 22$$

Substituting the value of $dx/dt$ from 10, we have:

$$\dot{V}(x(t), x_p(t), \Gamma) = -\left\| \nabla V(x(t), x_p(t), \Gamma) \right\|^2 \qquad 23$$

Since, from the maximum principle, V contains no local extrema in $\Omega$ (note that $\Gamma$, and $x_p \notin \Omega$) where, by design, the minimum is placed at $x_p$ and the maximum at $\Gamma$, the time derivative of the wave potential satisfies the conditions in 21. Therefore, V is a valid Liapunov Function. Hence, global asymptotic convergence to $x_p$ from anywhere in $\Omega$ is guaranteed.

In a similar way, it can be shown that for the singularity-free, dynamical system in 12 $dV/dt < 0$ $\forall x \in \Omega - B_\rho$, where

$$B_\rho(x): \{x: \|x - x_p\| < \rho\}. \qquad 24$$

This implies that:        $\lim_{t \to \infty} x(t) \in B_\rho(x).$           25

*D.1. Avoidance Analysis (the Dirichlet Case)*

Let $\Gamma\delta$ be a thin region surrounding $\Gamma$. Proving that motion will not be steered towards $\Gamma$ inside $\Gamma\delta$ is sufficient to prove that the forbidden regions will not be entered. Assuming that $x$ is initially inside $\Gamma\delta$, let $x_n$ be the distance between $x$ and $\Gamma$:

$$x_n = x^t \mathbf{n}, \qquad 26$$

where $\mathbf{n}$ denotes a unit vector normal to $\Gamma$. Since $\mathbf{n}$ is not a function of time, the time derivative of $x_n$ is equal to:

$$\dot{x}_n = \dot{x}^t \mathbf{n} = -(1 + \frac{1}{\|\nabla V\|^2} \cdot \frac{\partial V}{\partial t}) \nabla V^t \mathbf{n} = -(1 + \frac{1}{\|\nabla V\|^2} \cdot \frac{\partial V}{\partial t}) \frac{\partial V}{\partial n}. \qquad 27$$

Since the state is initially assumed to be outside $\Gamma$ ($x_n > 0$), proving that a measure of the length of $x_n$ is always non-decreasing is sufficient to prove that the state will never enter the forbidden regions. Let Va be a measure of $x_n$:

$$V_a = x_n^2. \qquad 28$$

The time derivative of Va may be computed as:

$$\dot{V}_a = 2 x_n \dot{x} = -2 x_n (1 + \frac{1}{\|\nabla V\|^2} \frac{\partial V}{\partial t}) \frac{\partial V}{\partial n}. \qquad 29$$

Since the value of V on $\Gamma$ is constrained to a constant C,

$$\frac{\partial V}{\partial t} = 0 \qquad x \in \Gamma. \qquad 30$$

Since $x$ is assumed to be very close to $\Gamma$ ($x \in \Gamma\delta$), the following approximation may be constructed:

$$\frac{\partial V}{\partial t} = 0 \qquad x \in \Gamma\delta. \qquad 31$$

Also, from the maximum principle, we have

$$V(x_1, x_p, \Gamma) > V(x_{21}, x_p, \Gamma), \qquad x_1 \in \Gamma, x_2 \in \Gamma_\delta. \qquad 32$$

Therefore
$$\frac{\partial V}{\partial n} < 0 \qquad x \in \Gamma_\delta \qquad 33$$

Using the above results it can be seen that the value of the time derivative of the Liapunov function inside $\Gamma\delta$ is equal to

$$\dot{V} \approx -2 x_n \frac{\partial V}{\partial n} > 0 \qquad 34$$

Therefore, the guidance field will always push $x$ away from $\Gamma\delta$, steering it away from $\Gamma$. In other words, the forbidden regions will be avoided.

*D.2. Avoidance Analysis (the Neumann Case)*

In the Neumann case $\partial V/\partial n$ is set to zero at $\Gamma$. Substituting this in 28, we have:

$$\dot{V} = 0 \qquad x \in \Gamma \qquad 35$$

Therefore motion can't proceed towards $\Gamma$ and the forbidden regions will not be entered.

VII Simulation Results
The tracking and region avoidance capabilities of the wave potential approach are tested.

A. Passive Evaders:
The capabilities of three types of partial differential operators are assessed for generating a spatio-temporal potential field that may be used to intercept a target using passive maneuvers. The operators are:

1-The Lpalacian operator: $\qquad \nabla^2 V = 0$, $\qquad\qquad$ 36

applied in a quasi stationary manner with the ODE system:

$$\dot{x} = -\nabla V .\qquad\qquad 37$$

2-The Diffusion operator: $\qquad \nabla^2 V = \dfrac{1}{a^2}\dfrac{\partial V}{\partial t}\qquad\qquad$ 38

the same ODE system in 37 is used.

3- The suggested Wave approach.

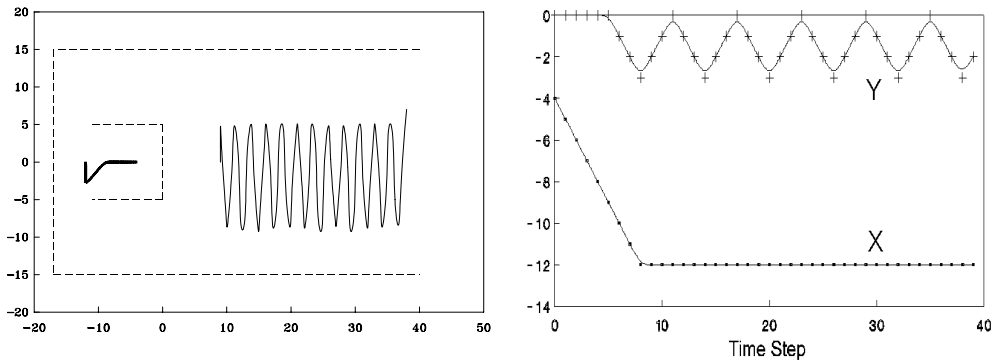

Figure-11: Response of the tracker, and X-Y components of trajectory Laplace.

The results from the three approaches are compared for a stationary, linearly moving, and slowly moving targets. It was observed that all techniques exhibit equivalent capabilities in terms of converging to the target and avoiding the forbidden regions. However, the disparity between the performances of the above techniques widened greatly when the linear motion of the target was augmented with rapid, sinusoidal oscillation. Figure-11 shows the response of the quasi stationary Laplace approach. As can be seen, the rapid fluctuations which the target superimposed along its escape path confused the tracker leaving it undecided whether to exit the corridor from the right side or the left side (figure-11). The total failure of the quasi-stationary Laplace strategy is a consequence of its total disregard to the temporal dependence of events and its reliance on spatial information only in guiding the tracker to the target. The absence of the time dimension from the strategy of the interceptor opens a "hatch" for the target through which this neglected dimension is exploited not only to evade capture by the interceptor, but even to totally paralyze it. Here, the target is aware that the interceptor is tracking it along the minimum distance path with no regard to where the target is moving to next, or how fast it is going there. Therefore, once the target observes a move of the tracker, it quickly repositions itself so that the distance of the interceptor to the new location of the target is shorter if the interceptor moves along a direction that is opposite to the one it is taking. By continuously repeating this relatively simple maneuver, the target is able to bring the motion of the interceptor to a standstill along the x-axis trapping it in a limit cycle along the y-axis.

Although the ODE system for the Diffusion strategy uses only spatial information to lay an interception trajectory for the target, temporal information is indirectly utilized in encoding the target and environment information in the potential field. In this case (figure-12), despite the ability of the tracker to proceed in the general direction of the target, it fails to keep up with its rapid fluctuating movements.

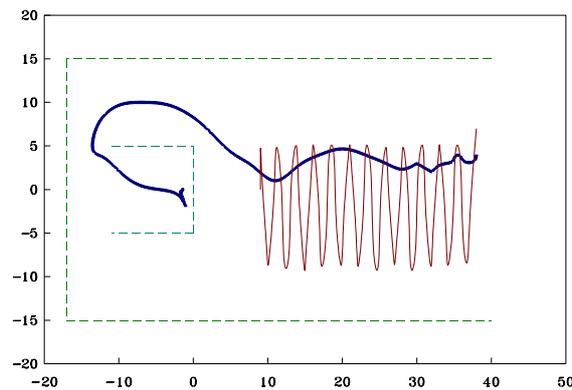
Figure-12: Response of the tracker, Diffusion .

As for the Wave Equation strategy, the interceptor was able to follow the target (figure-13).

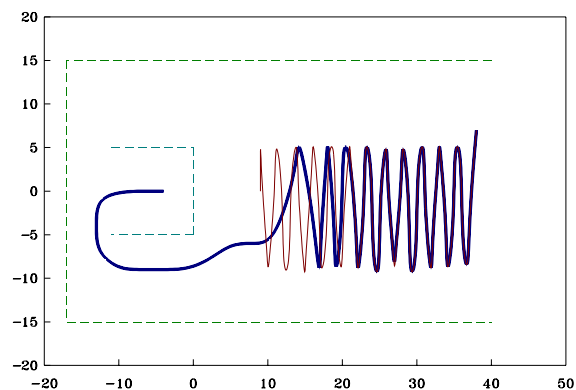
Figure-13: Response of the tracker, Wave.

B. Active Evaders:
Testing the wave approach with active evaders is more difficult than testing with the passive evader. The reason is the scarcity of techniques that enable an evader to plan its escape route in an environment that contains arbitrary clutter. In this paper, the technique used by the evader to play against a pursuer is based on the harmonic potential field approach which this author suggested in [19]. An equivalent approach that uses the Poisson potential may be found in [20]. Since the main objective of the dual (figure-4) is to compare the intelligence of the wave approach to that of the approach the evader is using, special attention has to be given to the notion intelligent behavior versus brute force behavior. Roughly speaking, one may consider the ability to provide a high magnitude of action as a brute force behavior characteristic. On the other hand, the process of projecting the effort along the right direction may be considered as the intelligent component of the planner. Therefore, a comparison of the methods used by the pursuer and the evader may be established by normalizing the guidance signal for each actor then scaling it with the speed each is using. For the pursuer we have:

$$\dot{X}_p = v_p \cdot \Phi_p, \qquad \Phi_p = \frac{g}{|g|} \qquad\qquad 39$$

and for the evader:

$$\dot{X}_e = v_e \cdot \Phi_e, \qquad \Phi_e = \frac{-\nabla V}{|\nabla V|} \qquad\qquad 40$$

where $X_p$ and $X_e$ are the x-y positions of the pursuer and the evader respectively. $v_p$, and $v_e$ are the speeds of the pursuer and the evader respectively.

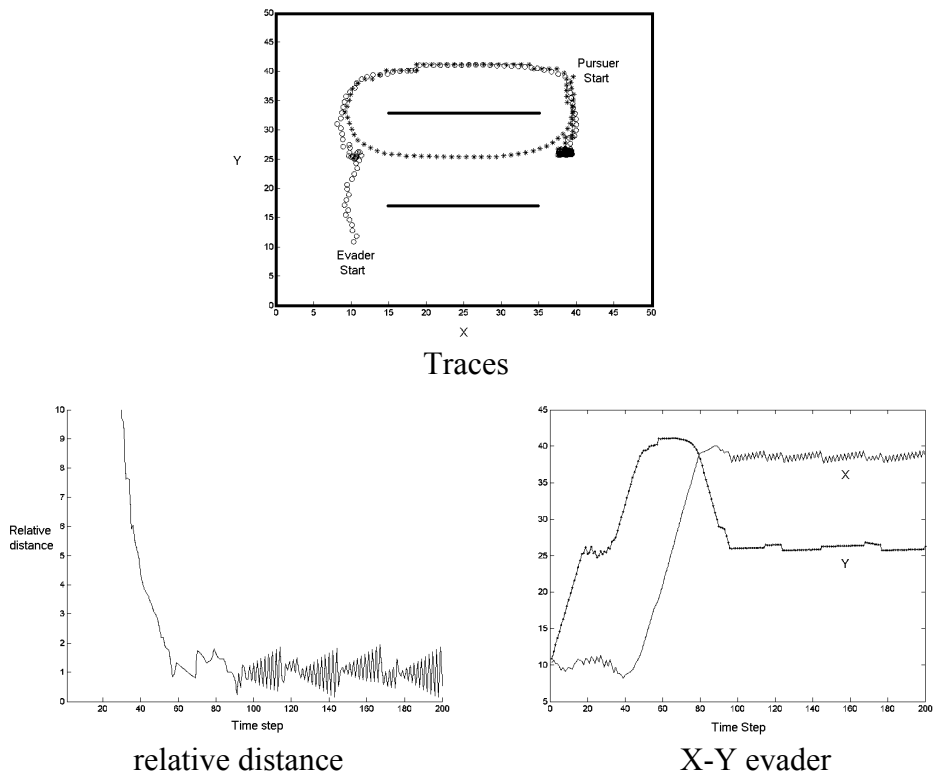

Traces

relative distance          X-Y evader

Figure-14: Evader-harmonic versus pursuer-harmonic, Ve=0.95, Vp=1.00

In figure-14 both the evader and the pursuer are using the harmonic approach to plan their movements against each other. The evader's speed is made to be 5% less than that of the pursuer. As can be seen, the pursuer managed to catch-up with the evader and pin it to a specific location in the workspace. In figure-15, the speed of the evader is set higher than that of the pursuer. As expected, the pursuer was not able to establish a lock on the evader which, most of the time, was able to maintain a four meter or less relative distance away from the pursuer. A four meter distance is considered to be the safe separation distance below which the evader which is using the harmonic approach begins to take escape action.

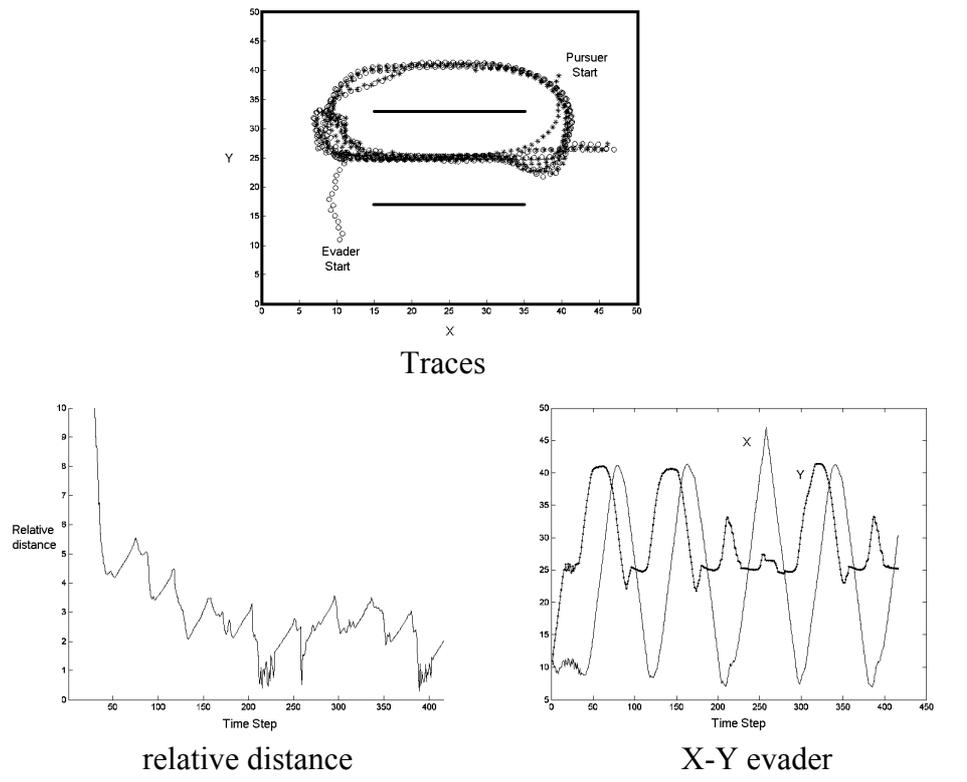

Traces

relative distance            X-Y evader

Figure-15: Evader-harmonic versus pursuer-harmonic, Ve=1.05, Vp=1.00.

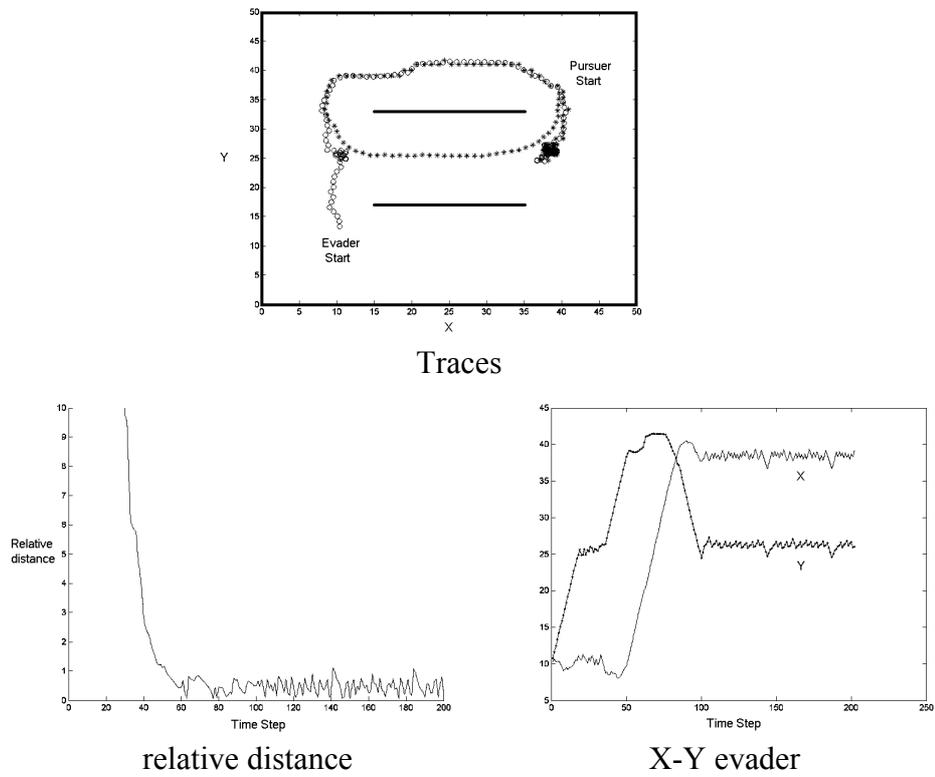

Traces

relative distance            X-Y evader

Figure-16: Evader-harmonic versus pursuer-wave, Ve=0.95, Vp=1.00

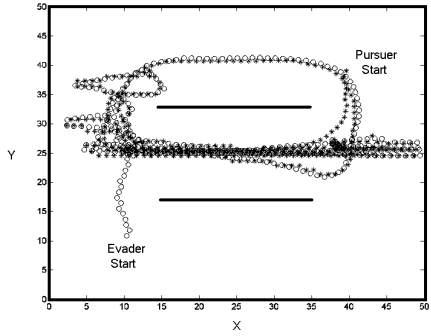

Traces

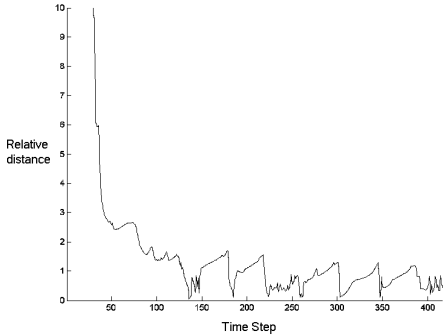

relative distance

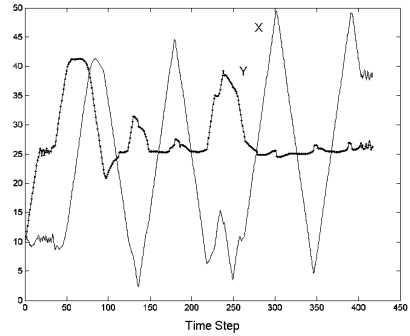

X-Y evader

Figure-17: Evader-harmonic versus pursuer-wave, Ve=1.05, Vp=1.00

In figure-16, the pursuer starts using the wave approach to plan its movements. The evader's speed is made 5% less than the speed of the pursuer. As expected, the pursuer managed to overcome the evader and pin it to a specific location in the workspace. It can be seen from the relative distance curve that the pursuer in this case was able to better lock onto the evader compared to the harmonic case. In figure-17, the speed of the evader is made 5% higher than that of the pursuer. Although in this case the pursuer was not able to lock the evader movement along the x-direction, it managed to stay very close to it maintaining a one and a half meter or less relative distance at all times.

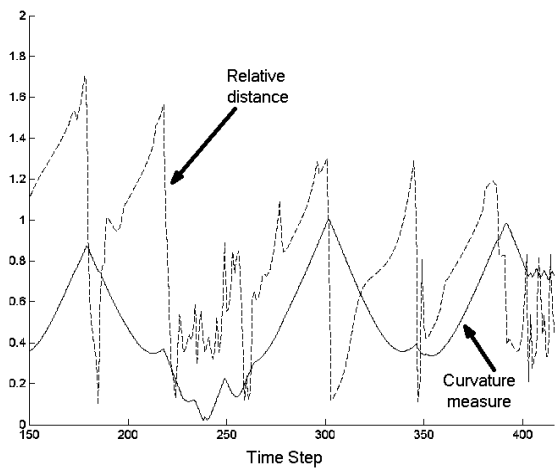

Figure-18: Sudden drop in relative distance coincides with peak curvature in the wave approach.

Figure-18 shows both a normalized curvature measure of the evader's trajectory and the relative distance between the pursuer and the evader for the case in figure-17. It is interesting to notice that a sudden drop in the relative distance always coincides with peaks in the curvature. This is a strong indication that the wave approach is doing a better job in directing motion than the harmonic approach. This observation could be an argument against the widely accepted protean escape maneuver [21,22]. In such a maneuver the evader keeps randomly changing the direction of the path, hence its curvature, in order to confuse the pursuer. However, figure-18 shows clearly that a pursuer guided by the wave approach exploits such fluctuation to make its gains in relative distance. This also confirms the widely accepted belief that the best policy for an evader with limited intellectual capabilities is to rely on brute force behavior by moving along a straight escape route (if it can) at the highest possible speed.

To test the ability of protean behavior as a credible means of evading intelligent pursuers, the harmonic potential evasion approach in [13] is adapted for use in random evasion. The adaption is based on the fact that a harmonic potential may be used to define a probabilistic collision measure of distance [5] from the evader's location to the pursuer and the forbidden regions. The block diagram of the random evasion planner is shown in figure-19. In the following simulation, a pursuer using the wave potential approach attempts to capture a random evader. The risk level the evader is operating at is 0.6. Its speed is made 20% higher than that of the pursuer. The trajectories of the pursuer and the evaders, the relative distance between the two and the x and y components of motion for the evader are shown in figure-20. As can be seen, the random evasion strategy was not able to make use of the considerable advantage in speed the evader has and the pursuer which is using the more intelligent wave strategy was able to catch up with the pursuer, even establish a lock on it.

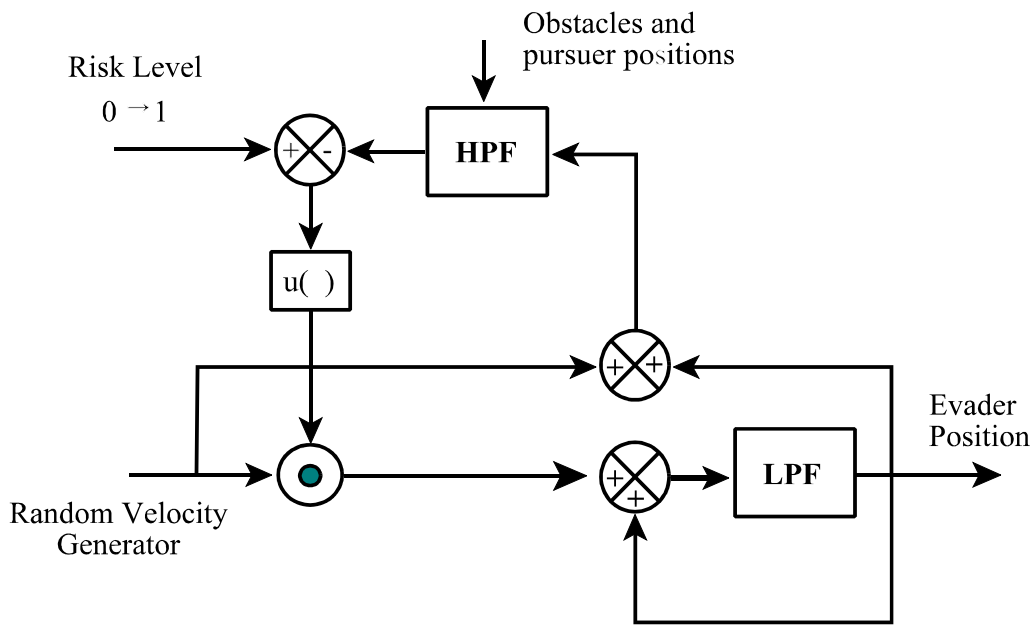

Figure 19: An HPF-based, random evasion strategy.

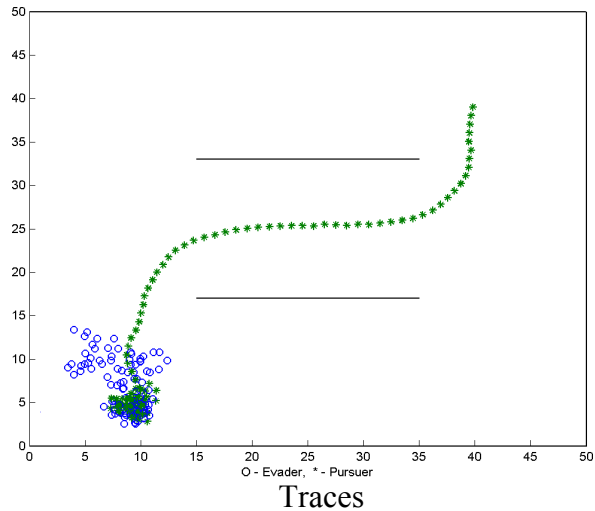

Traces

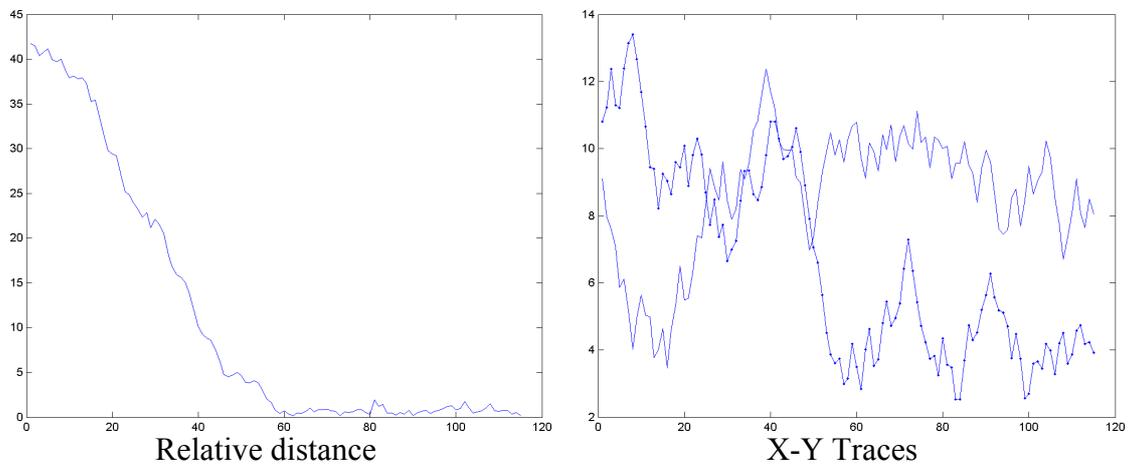

Relative distance

X-Y Traces

Figure 20: Evader-random versus pursuer-wave, ve = 1.20, vp

## VIII Conclusion

This paper focuses on a special class of pursuit-evasion problems concerned with intercepting a well-informed, active, intelligent target that is maneuvering to escape capture in a known, stationary, cluttered environment. The paper examines the possibility of the existence of a tracking method that is not dependent on the psychological profile of the prey or the manner it may utilize the knowledge it has to evade capture. It seems that the suggested wave potential-based controller has the ability to exhibit such objectivity in a goal-oriented, action synthesis. There are still many questions that need to be answered about the behavior of the suggested approach. For example, the ability of the tracker to always intercept the target regardless of the amount of information the target has, or the efficiency of its utilization needs to be carefully examined. Another question is concerned with the effect of placing dynamical constraints on the tracker's ability to intercept a target and the effect of delays caused by the potential field synthesis process on the interception ability of the pursuer. These are some of the important questions which future investigations of this approach will address.

**Acknowledgment**: The author thanks KFUPM for its support of this work.

**Appendix**

A. *Definition*: Let V(X) be a smooth ( at least twice differentiable), scalar function (V(X): $R^N$ → R). A point Xo is called a critical point of V if the gradient vanishes at that point ($\nabla V(Xo)=0$); otherwise, Xo is regular. A critical point is Morse, if its Hessian matrix (H(Xo)) is nonsingular. V(X) is Morse if all its critical points are Morse [23].

B. *Proposition*: If V(X) is a harmonic function defined in an N-dimensional space ($R^N$) on an open set $\Omega$, then the Hessian matrix at every critical point of V is nonsingular, i.e. V is Morse.

*Proof*: There are two properties of harmonic functions that are used in the proof:
1- a harmonic function (V(X)) defined on an open set $\Omega$ which contains no maxima or minima, local or global in $\Omega$. An extrema of V(X) can only occur at the boundary of $\Omega$,

2- if V(X) is constant in any open subset of $\Omega$, then it is constant for all $\Omega$.

Other properties of harmonic functions may be found in [10].

Let Xo be a critical point of V(X) inside $\Omega$. Since no maxima or minima of V exist inside $\Omega$, Xo has to be a saddle point. Let V(X) be represented in the neighborhood of Xo using a second order Taylor series expansion:

$$V(X) = V(Xo) + \nabla V(Xo)^T(X - Xo) + \frac{1}{2}(X - Xo)^T H(Xo)(X - Xo) \qquad |X-X0|<<1. \qquad 41$$

Since Xo is a critical point of V, we have:

$$V' = V(X) - V(Xo) = \frac{1}{2}(X - Xo)^T H(Xo)(X - Xo) \qquad |X-X0|<<1. \qquad 42$$

Notice that adding or subtracting a constant from a harmonic function yields another harmonic function, i.e. V` is also harmonic. Using eigenvalue decomposition:

$$V' = \frac{1}{2}(X-Xo)^T U^T \begin{bmatrix} \lambda_1 & 0 & 0 & 0 \\ 0 & \lambda_2 & . & 0 \\ . & . & . & . \\ 0 & 0 & . & \lambda_N \end{bmatrix} U(X-Xo) = \frac{1}{2}\xi^T \begin{bmatrix} \lambda_1 & 0 & 0 & 0 \\ 0 & \lambda_2 & . & 0 \\ . & . & . & . \\ 0 & 0 & . & \lambda_N \end{bmatrix} \xi = \frac{1}{2}\sum_{i=1}^{N} \lambda_i \xi_i^2 \qquad 43$$

where U is an orthonormal matrix of eigenvectors, $\lambda$'s are the eigenvalues of H(Xo), and $\xi=[\xi_1 \xi_2 ..\xi_N]^T = U(X-Xo)$. Since V` is harmonic, it cannot be zero on any open subset $\Omega$; otherwise, it will be zero for all $\Omega$, which is not the case. This can only be true if and only if all the $\lambda_i$'s are nonzero. In other words, the Hessian of V at a critical point Xo is nonsingular. This makes the harmonic function V also a Morse function.

It ought to be mentioned that a navigation function defined in [24] is a special case of a harmonic potential field. According to [24] a navigation function must satisfy the following properties:
1- it is smooth (at least $C^2$),
2- it contains only one minimum located at the target point,
3- it is a Morse function,
4- it is maximal and constant on $\Gamma$.

A harmonic function (V) is $C^\infty$ and Morse. Harmonic functions are extrema-free in $\Omega$. Their maxima and minima can only happen at the boundary of $\Omega$. In the harmonic approach $\Gamma$ and the target point ($X_T$) are treated as the boundary of Ω. Through applying the appropriate boundary conditions, the minimum of V is forced to occur on $X_T$. Also by the application of the Drichlet boundary conditions, the value of V is forced to be maximal and constant at Γ. The Drichlet condition (constant potential on the boundary) is one of many settings used in constructing a harmonic potential that may be used for navigation.